\documentclass{elsarticle}

\usepackage{amssymb}
\usepackage{multirow}
\usepackage{url}
\usepackage{lineno,hyperref}
\modulolinenumbers[5]

\journal{Neurocomputing}

\begin{document}

\begin{frontmatter}

\title{Cascade Region Proposal and Global Context for Deep Object Detection}

\author{Qiaoyong Zhong, Chao Li, Yingying Zhang, Di Xie, Shicai Yang, Shiliang Pu}
\address{Hikvision Research Institute, Hangzhou, China}
\ead{\{zhongqiaoyong,lichao15,zhangyingying7,xiedi,yangshicai,pushiliang\}@hikvision.com}

\begin{abstract}
  Deep region-based object detector consists of a region proposal step and
  a deep object recognition step. In this paper, we make significant
  improvements on both of the two steps. For region proposal we propose a
  novel lightweight cascade structure which can effectively improve RPN
  proposal quality. For object recognition we re-implement global context
  modeling with a few modifications and obtain a performance boost (4.2\%
  mAP gain on the ILSVRC 2016 validation set). Besides, we apply the idea
  of pre-training extensively and show its importance in both steps.
  Together with common training and testing tricks, we improve Faster
  R-CNN baseline by a large margin. In particular, we obtain 87.9\% mAP on
  the PASCAL VOC 2012 test set, 65.3\% on the ILSVRC 2016 test set and 36.8\%
  on the COCO test-std set.
\end{abstract}


\end{frontmatter}


\section{Introduction}

Object detection is a fundamental problem in computer vision. It has been
widely studied for many years~\cite{felzenszwalb2010object}, among which the
state-of-the-art approaches are based on convolutional neural networks
(CNN)~\cite{sermanet2013overfeat,simonyan2014very,erhan2014scalable,szegedy2014scalable,redmon2015you,liu2015ssd}.
Girshick et al.~\cite{girshick2014rich} proposed region-based CNN (R-CNN),
which successfully transfers the image-level recognition power of CNN to
object detection. Afterwards, R-CNN was subsequently developed and
accelerated in SPP-Net~\cite{he2015spatial}, NoC~\cite{ren2015object}, Fast
R-CNN~\cite{girshick2015fast} and Faster R-CNN~\cite{ren2015faster}.

In Faster R-CNN, a carefully designed region proposal network (RPN) was
introduced to extract high-quality region proposals. The extracted proposals
are then fed into Fast R-CNN (FRCN) for object recognition. In this paper,
we make extensive improvements on both region proposal quality (RPN side)
and object recognition accuracy (FRCN side). Our contribution is three-fold.
1) We propose a lightweight cascade RPN architecture, which can extract
accurately localized region proposals with marginal extra computational cost.
2) We revisit global context modeling. With a few modifications in network
architecture and the idea of pretraining, we obtain significant performance
gain. 3) We systematically evaluate common training and testing tricks that
can be found in the literature and report their contributions to detection
performance.

Based on these improvements over Faster R-CNN baseline, we achieve the
state-of-the-art performance on PASCAL VOC 2012~\cite{everingham2010pascal},
ILSVRC 2016~\cite{russakovsky2015imagenet} and COCO~\cite{lin2014microsoft}.

\section{Related Work}

\paragraph{Cascade Region Proposal} Conventional region proposal methods are
normally based on low-level features, either unsupervised (e.g. Selective
Search~\cite{uijlings2013selective} and EdgeBoxes~\cite{zitnick2014edge}) or
supervised (e.g. BING~\cite{cheng2014bing}). With the success of CNN in
computer vision~\cite{krizhevsky2012imagenet}, high-level semantic CNN
features are adopted for region proposal. Taking RPN as an example, a fully
convolutional architecture is designed to extract high-level features and
predict proposals in an end-to-end way. On the other hand, refining region
proposals with a multi-stage cascading pipeline has also been explored.
DeepBox~\cite{Kuo2015DeepBox} uses CNN to rerank proposals from a
conventional method, which may be considered as a special form of cascading.
CRAFT~\cite{yang2016craft} uses a two-class Fast R-CNN to refine proposals
from a standard RPN, which was further developed in~\cite{zeng2016crafting}.
DeepProposals~\cite{ghodrati2015deepproposal} proposes an inverse cascade to
exploit feature maps of different levels. Although our design shares similar
pipeline with CRAFT~\cite{yang2016craft}, we use a modified RPN instead of
Fast R-CNN in the second cascading stage. The details are described in
\ref{sec:cascade-rpn}.

\paragraph{Context Modeling} Objects in the real world do not exist on their
own. They are surrounded by a background (e.g. sky, grassland) and likely to
coexist with other objects, either of the same category or not. This context
may provide valuable information for discriminating objects of interest. In
R-CNN, classification of an object is solely based on the regional
information. In SPP-Net and Fast R-CNN, with the enlarged receptive fields
of convolution layers, contextual information is implicitly yet only
partially exploited as regional features are cropped from feature maps of
deep layers. Gidaris and Komodakis~\cite{gidaris2015object} proposed a
Multi-Region CNN architecture, where a rectangular ring around the object is
cropped and serves as context. \cite{chen20153d,cai2016unified} utilized an
enlarged area (e.g. 1.5$\times$) around the object as local context. Bell
et al.~\cite{bell16ion} used spatial recurrent neural networks (RNN) to model
context outside the object. Ouyang et al.~\cite{ouyang2015deepid} used
ImageNet 1000-class whole image classification scores as scene information
to refine per-object classification scores. He et al.~\cite{he2016deep}
proposed global context modeling, where the image-level features are
extracted through RoI pooling over the whole image. In this paper, we
reimplement global context modeling in \cite{he2016deep} with a few
modifications.

\section{Methods}

First of all, we will introduce the backbone network we use. Then our
improvements on both region proposal and classification are described,
followed by common training and testing tricks and our configurations.

\subsection{Backbone Network}

Following~\cite{he2016deep}, we choose ResNet-101 as the backbone network.
Since the birth of ResNet, it has been further developed. Gross and
Wilber~\cite{gross2016training} reimplemented ResNet with a few modifications,
e.g. down-sampling in the $3\times 3$ convolution instead of the first
$1\times 1$ convolution. Afterwards, He et al.~\cite{he2016identity} revisited
residual connection and proposed the identity mapping variant. In this paper,
we adopt both $3\times 3$ convolution down-sampling and identity mapping in
the ResNet-101 backbone network. The backbone network is pretrained on
the ImageNet classification dataset and then fine-tuned for detection tasks.

\subsection{Improving Region Proposal}

It is obvious and also has been reported~\cite{hosang2016makes} that
proposal quality is crucial for region based detectors. CNN-based RPN can
extract more compact and high-quality proposals than conventional methods.
Based on RPN, we propose several strategies to further improve proposal
quality, including pretraining RPN, cascade RPN and constrained ratio of
negative over positive anchors.

\subsubsection{Pretraining RPN} \label{sec:pretrain-rpn}

Pretraining on a large-scale dataset and transferring the learned features
to another task with smaller dataset have been the de facto strategy in deep
learning based applications. It has been widely reported that fine-tuning
from a pretrained model can combat overfitting
effectively~\cite{simonyan2014very}, leading to superior performance over
training from scratch. In this paper, we apply this idea extensively. The
basic principle is to \emph{pretrain as many layers as possible}.

\begin{figure}[!t]
\centering
\includegraphics[width=0.3\linewidth]{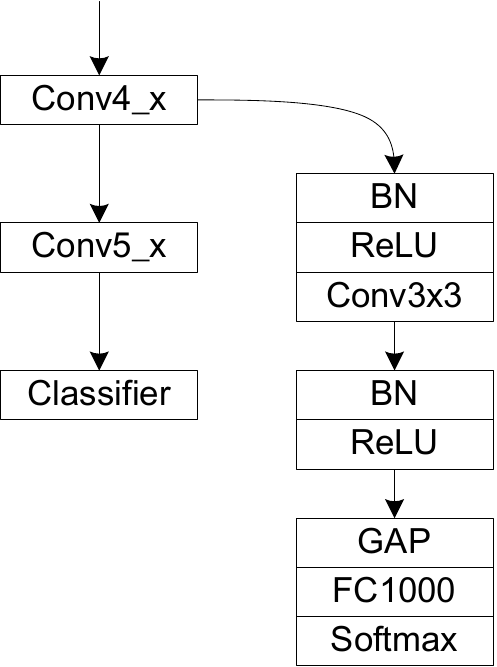}
\caption{Pretraining RPN with an auxiliary classifier. The trunk classifier
  on the left is for pretraining R-CNN, while the auxiliary classifier on the
  right is for pretraining RPN. GAP denotes global average pooling.}
\label{fig:pretrain-rpn}
\end{figure}

The RPN sub-network is connected to an intermediate layer, e.g. the last
layer of conv4\_x block for ResNets. And normally an extra $3\times 3$
convolution layer is added, which acts as a buffer to prevent direct
back-propagation of gradients from the RPN branch~\cite{cai2016unified}. In
Faster R-CNN baseline, this layer is randomly initialized and trained from
scratch during fine-tuning. We pretrain this layer using an auxiliary
classifier with a weight of 0.3 as in~\cite{szegedy2015going}.
Figure~\ref{fig:pretrain-rpn} shows the architecture. The preceding BN-ReLU
layers are inserted to comply with the identity mapping design. When
fine-tuning RPN, the linear classifier as well as global average pooling (GAP)
are replaced with RPN's classification and bounding box regression layers.

\subsubsection{Cascade RPN} \label{sec:cascade-rpn}
\begin{figure}[!t]
\centering
\includegraphics[width=0.8\linewidth]{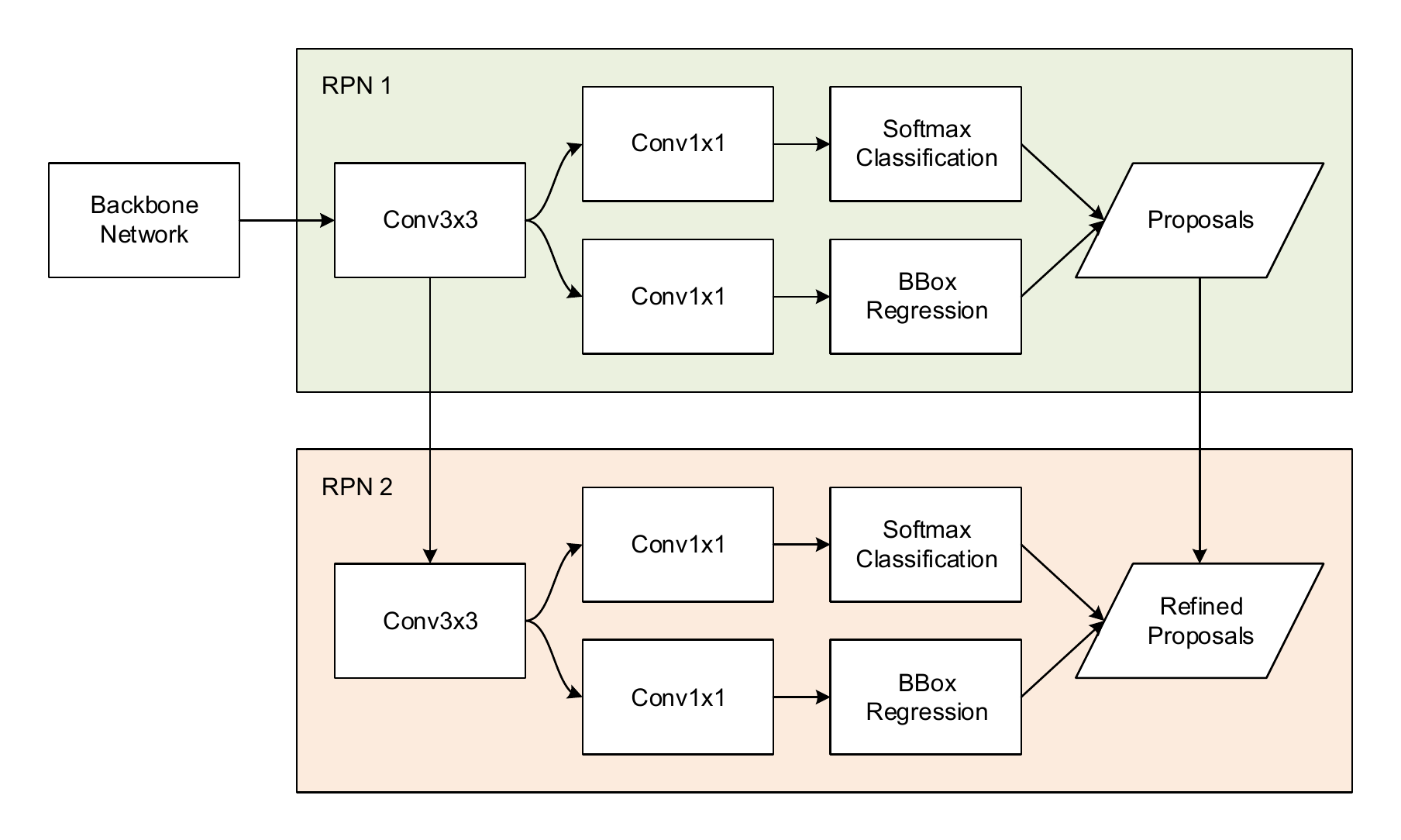}
\caption{Cascade RPN pipeline by stacking two RPN sub-networks sequentially.
  RPN 2 is adapted so as to take output proposals of RPN 1 as reference.}
\label{fig:cascade-rpn}
\end{figure}

We propose a lightweight cascade architecture to refine score and location
of RPN proposals. Figure~\ref{fig:cascade-rpn} shows the pipeline. The upper
part is a standard RPN (RPN 1), which utilizes sliding window proposals as
anchors and produces bounding box regressed proposals. Besides RPN 1 we add
another RPN sub-network (RPN 2), which takes output proposals of RPN 1 as
input. Note that the input proposals of RPN 2 refer to those right after
bounding box regression without any post-processing (e.g. sorting, non-maximum
suppression and truncating in number). Thus there is an one-to-one
correspondence between sliding window anchors and input proposals of RPN 2.
When training RPN 2, the sliding window anchors are used to locate the pixel
position in the feature map, while their corresponding proposals are used to
compute classification and bounding box regression targets.

During inference, we find that RPN 2 improves recall of medium and large
objects while hurts recall of small objects. So we combine small bounding
boxes of RPN 1 and medium-large bounding boxes of RPN 2 as the final set of
proposals. In our experiments, we set the size threshold to $64^2$ in pixels.

Compared with other cascade region proposal methods~\cite{yang2016craft},
our approach has the following advantages.

\paragraph{Easy to implement} The logic of RPN 2 is very similar to standard
RPN. We only need to replace the sliding window anchors with proposals from
RPN 1. The network architecture is rather brief and can be easily configured
with common deep learning frameworks like Caffe~\cite{jia2014caffe}.

\paragraph{Computationally efficient} \cite{yang2016craft} uses an extra
two-class Fast R-CNN to refine RPN proposals, which is computationally
inefficient. While our method works by stacking two RPN networks
sequentially. The fully convolutional nature makes it very efficient in
terms of both computational and memory cost.

\subsubsection{Constrained N/P Anchor Ratio}

Another improvement to RPN we make is to control (normally limit) the ratio
of negative and positive (N/P) anchors during training phase. In na\"{\i}ve
RPN, a common choice for anchor batch size is 256 and expected N/P ratio is
1. However in practice we find that this ratio may be very large, usually
greater than 10. The imbalance may lead to a bias towards background class,
thus hurting proposal recall. To address this issue, we add two more
hyper-parameters for training RPN, i.e. \texttt{max\_np\_ratio} and
\texttt{min\_batch\_size}. \texttt{max\_np\_ratio} works by shrinking batch
size when there are too few positive anchors. And \texttt{min\_batch\_size}
makes sure that the effective batch size will not become too small. In our
experiments, we arbitrarily set \texttt{max\_np\_ratio} to 1.5 and
\texttt{min\_batch\_size} to 32.

\subsection{Augmenting Classification with Pretrained Global Context}
\begin{figure}[!t]
\centering
\includegraphics[width=0.5\linewidth]{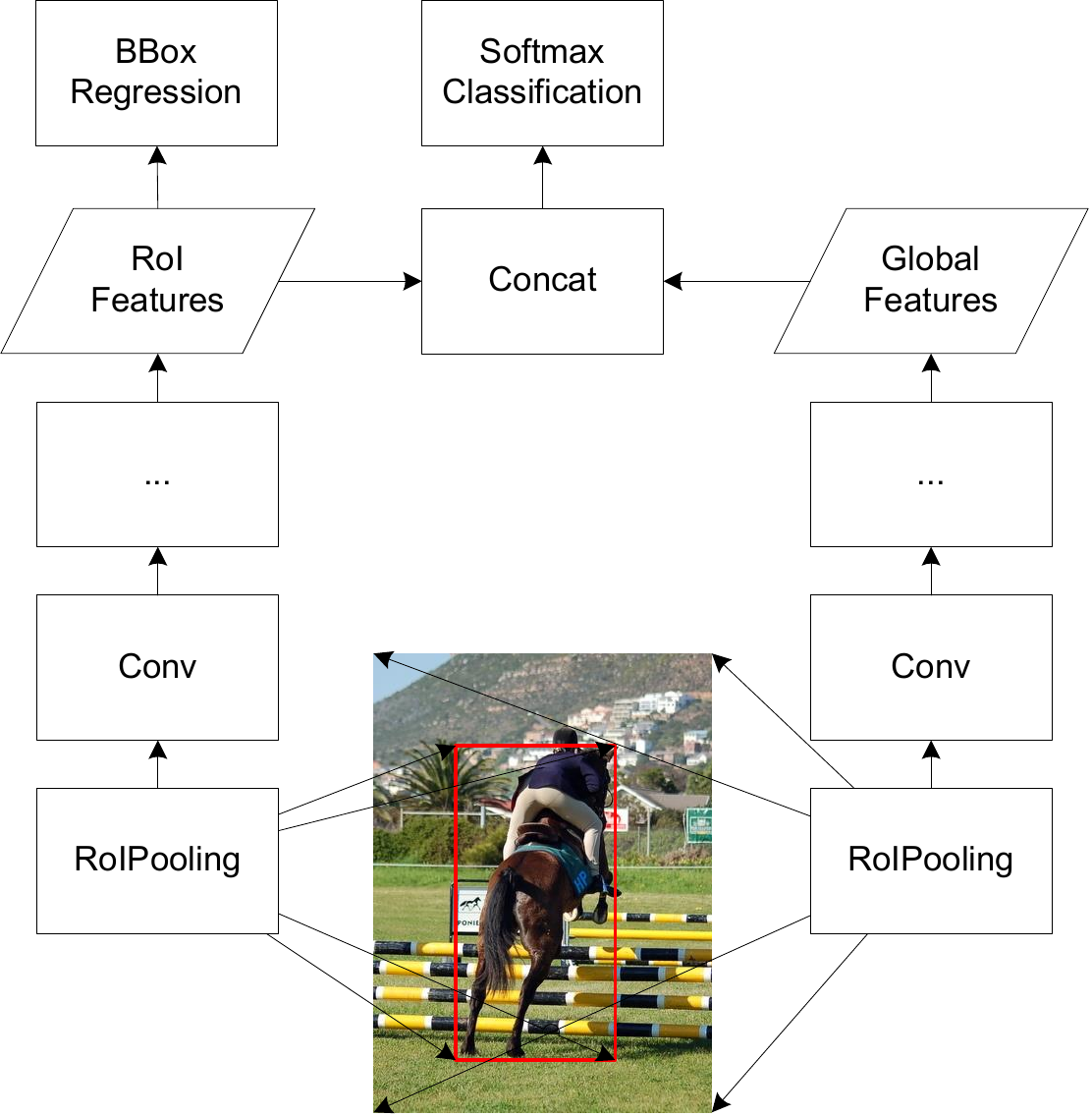}
\caption{Global context modeling through RoI pooling over the entire image.
  The combination of per-RoI features and global features are used to classify
  the objects of interest.}
\label{fig:global-context}
\end{figure}

Our implementation of global context is based on the work by He
et al.~\cite{he2016deep} with a few modifications.
Figure~\ref{fig:global-context} demonstrates our design. Besides the RoI
pooling over each RoI region, an RoI pooling over the entire image is
performed to extract global features. Since bounding box regression is based
on relative position and scale shift of target objects over proposals, the
surrounding context information may not help at all, or even confuses the
regressor. Thus unlike~\cite{he2016deep}, we use global context features for
classification only, not for bounding box regression.

Furthermore, we apply the pretraining principle
(section~\ref{sec:pretrain-rpn}) again on global context. For extremely deep
networks, the number of newly-added layers for the global context branch may
get quite large, e.g. 9 convolution layers for ResNet-101. When fine-tuning a
detection network, we normally choose a relative low base learning rate (e.g.
0.001). It would be difficult to train 9 layers from scratch with a low
learning rate. Pretraining is extremely critical in this scenario. In our
experiments, we simply copy the pretrained parameters of conv5\_x to the
global context branch for initialization.

\subsection{Training Tricks}
\subsubsection{Balanced Sampling}

\begin{figure}[!t]
\centering
\includegraphics[width=0.6\linewidth]{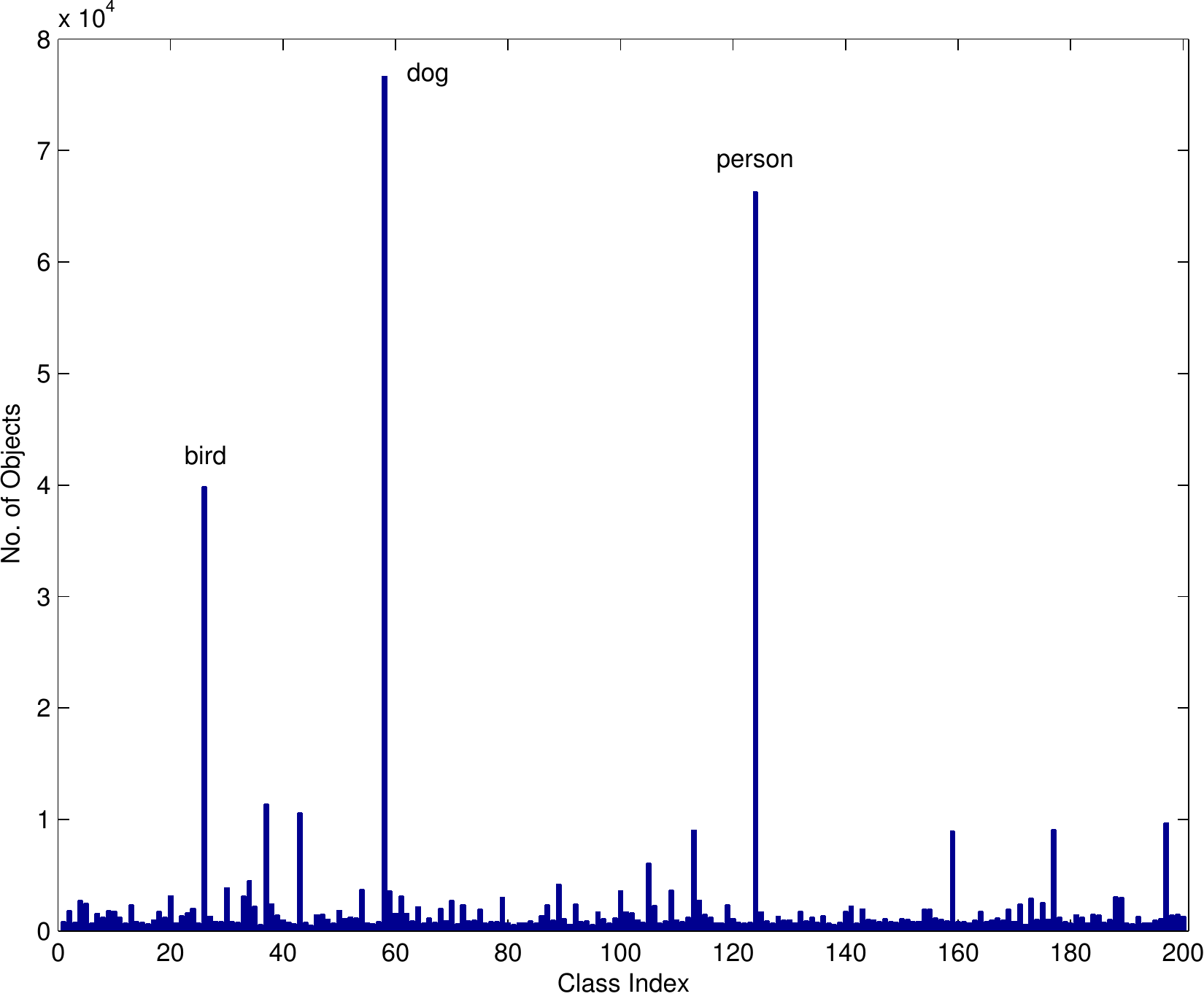}
\caption{Highly unbalanced distribution of the 200-class objects in the
  ImageNet DET training set. The top-3 frequent objects are dog, person and
  bird.}
\label{fig:det-class-distro}
\end{figure}

Data imbalance is a commonly occurring problem in vision recognition tasks.
For example, the ImageNet detection (DET) training set is highly unbalanced
(Figure~\ref{fig:det-class-distro}). The top-3 frequent classes (i.e. dog,
person and bird) comprise 36\% of all object instances. The biggest class
can be over 100 times larger than the smallest class. The conventional
sampling strategy is to construct an universal sample list for all classes
and read samples consecutively from the list, which will cause a bias
towards big classes.

Shen et al.~\cite{shen2016relay} proposed a class-aware sampling strategy to
cope with the imbalance issue in the scene classification task. We adapt
this strategy for object detection task. During training, we first sample
a class, then sample an image containing objects of this class.

\subsubsection{Negative Samples Training}
In the original implementation of Faster R-CNN, background images without
any object instance are discarded. In our opinion, these negative samples
are valuable in the sense that they enrich the background class and help
reduce false alarms. In our implementation, negative samples can be utilized
for training the classifier. While bounding box regression is not involved
as there is no object instance available.

\subsubsection{Multi-Scale Training}
For data augmentation, besides random horizontal flipping, we also adopt
multi-scale training. During training, the short side of the image is
resized to one of four scales \{400, 500, 600, 700\} randomly.

\subsubsection{Online Hard Example Mining}
We follow the online hard example mining (OHEM)
paper~\cite{shrivastava2016training} for training the FRCN branch. 300
proposals are fed to forward propagation, while the most difficult 64
proposals measured by loss value are selected for backward propagation. We
adopt the implementation by Dai et al.~\cite{dai2016r}.

\subsection{Testing Tricks}
During inference, common tricks like multi-scale testing, horizontal
flipping and weighted box voting are applied. Based on validation, we choose
the following hyper-parameter settings.
\begin{itemize}
  \item Multi-scale testing is used for both RPN and FRCN, i.e. three
      scales (\{400, 600, 800\}) for RPN, and five scales (\{200, 400,
      600, 800, 1000\}) for FRCN.
  \item Horizontal flipping is also used for both RPN and FRCN.
  \item Weighted box voting is used only for FRCN, as we find that it
      hurts RPN proposal recall at high IOU thresholds (e.g. 0.7).
\end{itemize}

When doing multi-scale testing and horizontal flipping, we use different
merging strategies for RPN and FRCN. For RPN, all sets of proposals from
various scale and flip settings are combined and their union set is
post-processed through non-maximum suppression (NMS). For FRCN, scores and
refined bounding boxes of various scale and flip settings are averaged.

Furthermore, IOU thresholds of NMS are tuned for different datasets based on
performance on validation set. Specifically, we use 0.4 on ImageNet and 0.45
on PASCAL VOC and COCO instead of the commonly used 0.3.

\subsection{Implementation Details}
Based on
\emph{py-faster-rcnn}\footnote{\url{https://github.com/rbgirshick/py-faster-rcnn}},
we reimplement Faster R-CNN in C++ under the Caffe
framework~\cite{jia2014caffe}, which allows us to perform efficient
multi-GPU training. We adopt joint training of RPN and FRCN instead of the
4-step training scheme for convenience. In all experiments, we use the
following settings.

\begin{itemize}
  \item The RPN anchor scales are extended to 6 scales (\{32, 64, 96, 128,
      256, 512\}).
  \item Per-GPU RPN batch size is unchanged (256), while FRCN batch size
      is reduced from 128 to 64.
  \item Models are trained using 4 GPUs. Thus the effective batch size is
      4.
  \item The conventional SGD with 0.9 momentum, 0.0001 weight decay and
      the \emph{step} learning rate policy with 0.001 base learning rate
      are used for optimization.
\end{itemize}

\section{Results and Discussion}

To evaluate performance of our approach, as well as common training and
testing tricks, we perform extensive ablation studies on ImageNet detection
(DET) and localization (LOC) datasets. Besides, results on PASCAL VOC 2012 and
COCO are also reported.

\subsection{ImageNet DET}

Following~\cite{girshick2014rich}, we split the validation set into two
parts (val1 and val2). train+val1 set is used for training, and val2 set is
used for parameter tuning and model selection. The total number of training
samples is $342,679$ for positive samples only and $466,454$ including the
extra negative training data.

In each experiment, we train the model for 200k iterations with the learning
rate of 0.001, and then for extra 100k iterations with 0.0001.

\subsubsection{Faster R-CNN Baseline} \label{sec:faster-rcnn-baseline}

We pretrain the ResNet-101 backbone network on the ImageNet classification
dataset. The top-1 accuracy on the validation set is 76.3\%, which is on par
with~\cite{he2016deep}. With fine-tuning on the ImageNet detection dataset,
we obtain a baseline Faster R-CNN model, whose mAP is 52.6\% on val2.

\subsubsection{Improved Region Proposal}\label{sec:improved-proposals}
Based on the Faster R-CNN baseline, ablation studies are performed on region
proposal improvements. Table~\ref{tab:improve-region-proposals-result} shows
the results. Compared with baseline, pretrained RPN improves recall@0.5 by 1.2
points and mAP@0.5 by 0.3 point. While no improvement on recall at high IOU
thresholds is obtained. Cascade RPN improves recall@0.7 by 8.8 points and
average recall (AR) by 4.8 points, leading to 0.8\% mAP@0.7 gain. Constrained
N/P further increases recall by up to 3 points. Combining all the three
strategies, we obtain 2\% recall@0.5 gain, 11.2\% recall@0.7 gain and 6.9\% AR
gain. In terms of detection performance, we obtain 0.5\% mAP gain at normal
IOU threshold.

From the results we can conclude that our approach significantly improve
localization accuracy of region proposals. However, mAP gain is relatively
limited. After the competition, we have more time to look into this issue.
The reason might be that the extra RPN branch of our cascade RPN architecture
introduces too much impact on back-propagation of region proposal errors. In
other words, as RPN and FRCN are jointly trained within one network, FRCN
performance declines as RPN gets augmented. To validate our hypothesis, we
design two more experiments. In the first experiment, we reduce loss weight of
RPN by half, i.e. from 1 to 0.5 (LW=0.5). In the second experiment, we replace
joint training with a 2-step training scheme, in which RPN and FRCN are
optimized independently. In this case, the possible competition between RPN
and FRCN is fully eliminated.

The results are displayed in Table~\ref{tab:improve-region-proposals-result}.
After reducing loss weight of RPN, we obtain an extra 0.9\% and 1.2\% mAP gain
at 0.5 and 0.7 IOU thresholds respectively, even though proposal recall
declines slightly. When using the 2-step training scheme, proposal recall gets
improved, while detection mAP remains on par the reduced loss weight setting.
From these experiments we may conclude that while joint training of RPN and
FRCN can achieve comparable performance to separate training, balancing the
loss weights of RPN and FRCN is very important. With the issue addressed, in
total we obtain 1.4\% mAP gain at 0.5 IOU and 1.9\% at 0.7 IOU from region
proposal improvements.

On an M40 graphics card, the average running time over 1000 samples to
extract 300 proposals is 87ms for the RPN baseline and 98ms for our cascade
RPN. The extra computational cost (11ms) is marginal, considering the
significant improvement on proposal quality and detection performance.

\begin{table}[!t]
\renewcommand{\arraystretch}{1.2}
\centering
\begin{tabular}{l|cccc|cc}
\hline
Pretrained RPN & & \checkmark & \checkmark & \checkmark & \checkmark & \checkmark \\
Cascade RPN & & & \checkmark & \checkmark & \checkmark & \checkmark \\
Constrained N/P & & & & \checkmark & \checkmark & \checkmark \\
\hline
LW=0.5 & & & & & \checkmark &  \\
2-step training & & & & & & \checkmark \\
\hline
Recall@0.5 & 88.5 & 89.7 & 88.5 & 90.5 & 90.0 & \textbf{90.7} \\
Recall@0.7 & 67.1 & 66.5 & 75.3 & 78.3 & 77.4 & \textbf{78.7} \\
AR@[0.5:0.95] & 49.9 & 50.0 & 54.8 & 56.8 & 56.0 & \textbf{58.0} \\
\hline
mAP@0.5 & 52.6 & 52.9 & 52.9 & 53.1 & \textbf{54.0} & \textbf{54.0} \\
mAP@0.7 & 39.1 & 39.3 & 40.1 & 39.8 & \textbf{41.0} & 40.9 \\

\hline
\end{tabular}
\caption{Region proposal recalls and detection results on the ImageNet DET
 val2 set. All results are based on 300 proposals. AR denotes average
 recall over IOU thresholds from 0.5 to 0.95 with a step of 0.05.}
\label{tab:improve-region-proposals-result}
\end{table}

\subsubsection{Pretrained Global Context}

Based on current best model (not including the post-competition improvement)
with improved region proposal, we design two experiments on global context. In
the first experiment, the global context branch in
Figure~\ref{fig:global-context} is randomly initialized with the \emph{xavier}
policy~\cite{glorot2010understanding} and trained from scratch. While in the
second experiment, it is fine-tuned from pretrained parameters. The results
are shown in Table~\ref{tab:global-context-result}.  In the case of
pretraining, global context works surprisingly well, with a 4.2 points mAP
gain. While nearly no improvement is obtained in the case of random
initialization. \cite{he2016deep} reported a 1 point mAP gain with global
context on COCO dataset~\cite{lin2014microsoft}. Although it is not comparable
across different datasets, our implementation is proven to be very effective.

\begin{table}[!t]
\renewcommand{\arraystretch}{1.2}
\centering
\begin{tabular}{l|c|c|c}
\hline
 & Baseline & Random & Pretrained \\
\hline
mAP & 53.1 & 53.2 & \textbf{57.3} \\
\hline
\end{tabular}
\caption{Detection results of global context modeling on the ImageNet DET
  val2 set. Baseline: no global context. Random: randomly initialized
  global context. Pretrained: pretrained global context.}
\label{tab:global-context-result}
\end{table}

\subsubsection{Training and Testing Tricks}

Taking the network with pretrained global context as the new baseline, we
report further performance improvements from common training and testing
tricks in Table~\ref{tab:train-test-tricks-result}. Balanced sampling leads
to 2.5 points mAP gain, which is reasonable considering the severe data
imbalance. With negative samples added for training mAP increases by 0.8
point. While no improvement is observed using multi-scale training. By OHEM,
an extra 1.2 points mAP gain is obtained, leading to 61.7\% mAP without
testing tricks that may increase running time dramatically. With multi-scale
testing, horizontal flipping and weighted box voting, mAP reaches 64.1\%.
With pretraining on the ImageNet localization (LOC) dataset, an extra 0.8
point mAP gain is obtained. Finally, by combing the best region proposals
and object classifier, mAP reaches 65.1\%, which is our best single-model
result.

\begin{table*}[!t]
\renewcommand{\arraystretch}{1.2}
\setlength{\tabcolsep}{3pt}
\centering
\small
\begin{tabular}{l|l|c|c|c|c|c|c|c|c|c|c}
\hline
\multirow{4}{*}{Train} & Balanced sampling & & \checkmark & \checkmark & \checkmark & \checkmark & \checkmark & \checkmark & \checkmark & \checkmark & \checkmark \\
& +Negative samples & & & \checkmark & \checkmark & \checkmark & \checkmark & \checkmark & \checkmark & \checkmark & \checkmark \\
& Multi-scale training & & & & \checkmark & \checkmark & \checkmark & \checkmark & \checkmark & \checkmark & \checkmark \\
& OHEM & & & & & \checkmark & \checkmark & \checkmark & \checkmark & \checkmark & \checkmark \\
\hline
\multirow{3}{*}{Test} & Multi-scale testing & & & & & & \checkmark & \checkmark & \checkmark & \checkmark & \checkmark \\
& Horizontal flipping & & & & & & & \checkmark & \checkmark & \checkmark & \checkmark \\
& Box voting & & & & & & & & \checkmark & \checkmark & \checkmark \\
\hline
\multicolumn{2}{c|}{Pretrain on LOC} & & & & & & & & & \checkmark & \checkmark \\
\multicolumn{2}{c|}{Comb. best RPN \& FRCN} & & & & & & & & & & \checkmark \\
\hline
\multicolumn{2}{c|}{mAP} & 57.3 & 59.8 & 60.6 & 60.5 & \textbf{61.7} & 63.3 & 63.9 & 64.1 & 64.9 & \textbf{65.1} \\
\hline
\end{tabular}
\caption{Detection results of common training and testing tricks on the
 ImageNet DET val2 set.}
\label{tab:train-test-tricks-result}
\end{table*}

\subsubsection{Model Ensemble}

With model ensemble, we obtain 67.0\% mAP on the ImageNet DET val2 set and
65.3\% mAP on the test set (Table~\ref{tab:ilsvrc2016-result}), which ranks
2nd in ILSVRC 2016 challenge. In terms of single-model result, we obtain
63.4\% mAP on the test set, surpassing CUImage~\cite{zeng2016crafting} with
slight advantage. Compared with the MSRA baseline~\cite{he2016deep}, we
improve mAP by 4.6 points for single-model result and 3.2 points for ensemble.

\begin{table}[!t]
\renewcommand{\arraystretch}{1.2}
\centering
\begin{tabular}{l|l|c|c}
\hline
\multicolumn{2}{c|}{}  & val2 & test \\
\hline
\multirow{3}{*}{Single model} & MSRA~\cite{he2016deep} (ILSVRC'15)    & 60.5 & 58.8  \\
\cline{2-4}
 & Ours (ILSVRC'16)    & \textbf{65.1} & \textbf{63.4}  \\
 & CUImage~\cite{zeng2016crafting} (ILSVRC'16) & 65.0   & 63.36 \\
\hline
\multirow{3}{*}{Ensemble} & MSRA~\cite{he2016deep} (ILSVRC'15)    & 63.6 & 62.1  \\
\cline{2-4}
 & CUImage~\cite{zeng2016crafting} (ILSVRC'16) & \textbf{68.0}   & \textbf{66.3}  \\
 & Ours (ILSVRC'16)    & 67.0   & 65.3  \\
\hline
\end{tabular}
\caption{Final detection results on the ImageNet DET val2 and test set.}
\label{tab:ilsvrc2016-result}
\end{table}

\subsection{ImageNet LOC --- Localization by Detection}
The object localization task (LOC) of ILSVRC is more challenging as the
number of classes (1000) is much larger than DET (200). He
et al.~\cite{he2016deep} proposed a \emph{per-class} RPN + R-CNN pipeline for
object localization. On the contrary, we simply apply our object detection
approach to object localization. Since localization performance is limited
by classification performance, for a fair comparison, we use a
classification model with equivalent error rate as in~\cite{he2016deep}. As
shown in Table~\ref{tab:loc-result}, given the same classification error
rate (4.6\%), our localization error rate on the validation set is 10.2\%,
surpassing the MSRA baseline~\cite{he2016deep} by 0.4 point. By model
ensemble, we further reduce localization error rate to 8.8\%. On the test set,
even with a relatively inferior classification performance (3.7\% versus 3.6\%
top-5 error rate), our localization error rate reaches 8.7\%, surpassing the
MSRA baseline~\cite{he2016deep} by 0.3 point. Our submission won the 2nd place
in the object localization task of ILSVRC 2016.

\begin{table}[!t]
\renewcommand{\arraystretch}{1.2}
\setlength{\tabcolsep}{4pt}
\centering
\begin{tabular}{l|l|c|c|c|c}
\hline
\multicolumn{2}{c|}{} & \multicolumn{2}{c|}{val} & \multicolumn{2}{c}{test} \\
\cline{3-6}
\multicolumn{2}{c|}{} & LOC err. & CLS err. & LOC err. & CLS err. \\
\hline
\multirow{2}{*}{Single model} & MSRA~\cite{he2016deep} & 10.6 & 4.6 & - & - \\
& Ours & \textbf{10.2} & 4.6 & - & - \\
\hline
\multirow{2}{*}{Ensemble} & MSRA~\cite{he2016deep} & 8.9 & - & 9.0 & 3.6 \\
  & Ours & \textbf{8.8} & 3.5 & \textbf{8.7} & 3.7 \\
\hline
\end{tabular}
\caption{Object localization results (top-5 error) on the ImageNet LOC
  validation and test set.}
\label{tab:loc-result}
\end{table}

\subsection{PASCAL VOC}
Besides ImageNet, we also evaluate our approach on PASCAL VOC 2012. For VOC
data, we use trainvaltest set of VOC 2007 as well as train set of VOC 2012
(07++12) for training and val set of VOC 2012 for validation. COCO dataset is
used as extra training data (COCO+07++12). Since COCO has 80 categories, we
only use those 20 categories that are present in PASCAL VOC. After training on
COCO+07++12 converges, we further fine-tune the model on VOC 07++12.

Table~\ref{tab:voc2012-result} shows the detection results on both val and
test sets of PASCAL VOC 2012. Using standard PASCAL VOC training data, our mAP
is 82.9\% on val set and 81.8\% on test set. With extra COCO data, mAP is
boosted to 86.7\% and 86.0\% respectively, surpassing the MSRA
baseline~\cite{he2016deep} by 2.2 points. This performance is still
competitive to more recent methods like Deformable R-FCN~\cite{dai17dcn} and
ResNeXt-101~\cite{xie2017}, considering the fact that they use the val set of
VOC 2012 as extra training data while we do not. With further pretraining on
ImageNet DET, our mAP on test set reaches \textbf{87.9}\%, which is the new
state-of-the-art single-model performance at the time of writing this paper.

\begin{table}[!t]
\renewcommand{\arraystretch}{1.2}
\centering
\begin{tabular}{l|l|c|c}
\hline
\multirow{2}{*}{Method} & \multirow{2}{*}{Training data} & \multicolumn{2}{c}{mAP} \\
\cline{3-4}
& & val & test \\
\hline
MSRA~\cite{he2016deep} & COCO+07++12 & - & 83.8 \\
Deformable R-FCN~\cite{dai17dcn} & COCO+07++12 & - & 87.1 \\
ResNeXt-101 & COCO+07++12 & - & 86.1 \\
R-FCN~\cite{dai2016r} & COCO+07++12 & - & 85.0 \\
\hline
Ours & 07++12$^\ast$ & 82.9 & 81.8$^\dag$ \\
Ours & COCO+07++12$^\ast$ & 86.7 & 86.0$^\ddag$ \\
Ours & DET+COCO+07++12$^\ast$ & \textbf{88.7} & \textbf{87.9}$^\S$ \\
\hline
\end{tabular}
\caption{\textbf{Single-model} detection results on PASCAL VOC 2012.
$^\ast$Note that unlike other submissions, val set of VOC 2012 is not used for
training in all of our experiments.
  \normalfont $^\dag$: \protect\url{http://host.robots.ox.ac.uk:8080/anonymous/LQBOCC.html}.
  $^\ddag$: \protect\url{http://host.robots.ox.ac.uk:8080/anonymous/TK9SNW.html}.
  $^\S$:
  \protect\url{http://host.robots.ox.ac.uk:8080/anonymous/HHS4NF.html}.}
\label{tab:voc2012-result}
\end{table}

\subsection{COCO}
\begin{figure*}[!t]
\centering
\includegraphics[width=1.05\linewidth]{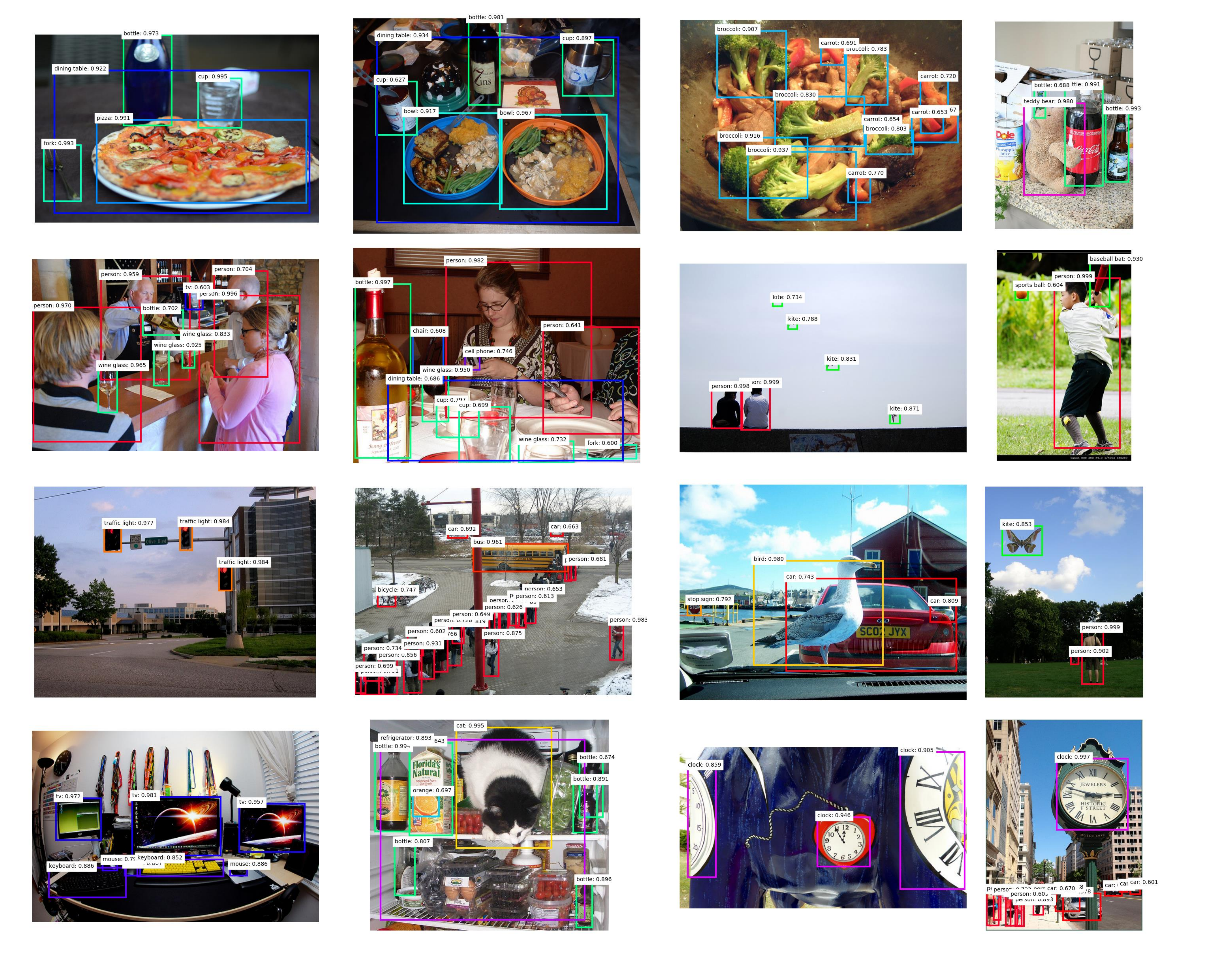}
\caption{Examples of our detection results on the COCO test-dev set. Best
  viewed in color.}
\label{fig:coco-examples}
\end{figure*}

We further evaluate our approach on the more challenging COCO dataset.
Table~\ref{tab:coco-result} shows a comparison of the current
state-of-the-arts in terms of single-model performance on the test-dev and
test-std sets. On test-dev, we achieve 36.7\% AP, surpassing the
MSRA~\cite{he2016deep} baseline by 1.8 points. In particular, our approach is
superior over the very recent FPN~\cite{lin2016fpn}, where feature maps of
multiple scales are exploited. Deformable R-FCN~\cite{dai17dcn} is the current
leading method. However, they use Inception-ResNet, which is a much stronger
backbone network than ResNet-101 we use. At 0.5 IOU threshold, we get 60.3\%
and 60.5\% AP on test-dev and test-std respectively, which is significantly
superior over all other approaches. This is due to our improvements (e.g.
pretrained global context) on FRCN object classification accuracy.
Figure~\ref{fig:coco-examples} shows some examples of our detection results.
Our approach is robust against variations of object size and aspect ratio.

\begin{table}[!t]
\renewcommand{\arraystretch}{1.2}
\centering
\begin{tabular}{l|l|c|c|c|c}
\hline
  \multirow{2}{*}{Method} & \multirow{2}{*}{Backbone} &
  \multicolumn{2}{c|}{test-dev} & \multicolumn{2}{c}{test-std} \\
\cline{3-6}
  & & AP & AP@.5 & AP & AP@.5\\
\hline
  Deformable R-FCN~\cite{dai17dcn} & Inception-ResNet & \textbf{37.5} & 58.0 &
  \textbf{37.3} & 58.0 \\
\hline
  Ours & ResNet-101 & 36.7 & \textbf{60.3} & 36.8 & \textbf{60.5} \\
\hline
  FPN~\cite{lin2016fpn} & ResNet-101 & 36.2 & 59.1 & 35.8 & 58.5 \\
\hline
  MSRA~\cite{he2016deep} & ResNet-101 & 34.9 & 55.7 & - & - \\
\hline
  G-RMI~\cite{Huang_2017_CVPR} & Inception-ResNet & 34.7 & - & - & - \\
\hline
\end{tabular}
\caption{\textbf{Single-model} detection results on COCO test-dev and test-std
  sets of the current state-of-the-arts.}
\label{tab:coco-result}
\end{table}

\section{Conclusions}
This paper proposes several improvements on both region proposal and region
classification, in particular the novel cascade RPN architecture and
improved global context modeling. Due to the computational efficiency, they
are applicable in practical scenarios (e.g. embedded platforms) where
computational resource is limited. Common training and testing tricks are
adopted and systematically evaluated in the context of large scale object
detection, which may ease choosing appropriate tricks given a fixed
computational budget. Our approach surpasses baseline by a significant
margin and achieves the state-of-the-art performance on PASCAL VOC, ILSVRC
2016 and COCO.

\section*{References}

\bibliography{main}

\end{document}